\documentclass{article}
\usepackage{spconf,amsmath,graphicx,hyperref}
\usepackage{booktabs}
\usepackage{multirow}
\usepackage{placeins} 

\title{MAGE-KT: Multi-Agent Graph-Enhanced Knowledge Tracing \\ with Subgraph Retrieval and Asymmetric Fusion}

\name{Chi Yu$^{1,\dagger}$ \qquad Hongyu Yuan$^{1,\dagger}$ \qquad Zhiyi Duan$^{1}$\sthanks{Corresponding author. $^{\dagger}$These authors contributed equally to this work.}}
\address{$^{1}$Department of Computer Science, Inner Mongolia University, Hohhot, China}

\begin{document}
\ninept
\maketitle
\begin{abstract}
Knowledge Tracing (KT) aims to model a student's learning trajectory and predict performance on the next question. 
A key challenge is how to better represent the relationships among students, questions, and knowledge concepts (KCs). 
Recently, graph-based KT paradigms have shown promise for this problem. 
However, existing methods have not sufficiently explored inter-concept relations, often inferred solely from interaction sequences.  
In addition, the scale and heterogeneity of KT graphs make full-graph encoding both computationally both costly and noise-prone, causing attention to bleed into student-irrelevant regions and degrading the fidelity of inter-KC relations. 
To address these issues, we propose a novel framework: \textbf{M}ulti-\textbf{A}gent \textbf{G}raph-\textbf{E}nhanced \textbf{K}nowledge \textbf{T}racing (MAGE-KT). 
It constructs a multi-view heterogeneous graph by combining a multi-agent KC relation extractor and a student–question interaction graph, capturing complementary semantic and behavioral signals.
Conditioned on the target student's history, it retrieves compact, high-value subgraphs and integrates them using an Asymmetric Cross-attention Fusion Module to enhance prediction while avoiding attention diffusion and irrelevant computation. 
Experiments on three widely used KT datasets show substantial improvements in KC-relation accuracy and clear gains in next-question prediction over existing methods.
\end{abstract}
\section{Introduction}
\label{sec:intro}

Knowledge tracing (KT) is a core task in educational data mining that infers a student's evolving knowledge state from past responses to forecast future performance \cite{Yudelson2013iBKT}. 
As a foundation of personalized learning, KT has been extensively explored, with substantial advances in modeling answer sequences \cite{ktsurvey}.

As Deep Learning (DL) has advanced, KT methods such as DKT \cite{Piech2015DKT}, DKT+ \cite{Yeung2018PCR}, and DKVMN \cite{Zhang2017DKVMN} have sparked a new wave of research, yet they lack explicit modeling of knowledge structure. 
To address this gap, graph-based approaches have been introduced to better capture relations among students, questions, and knowledge concepts (KCs). 
More recently, researchers have focused on relations within the KC space itself, examining interdependencies among concepts to more faithfully model real learning processes, as such relations have been shown to significantly boost KT performance. 
Existing techniques for extracting KC–KC relations fall into two categories: 
(i) inferring predecessor–successor and associative links from the temporal order in which KCs appear in students' interaction histories \cite{Tong2020SCKT, pan2024knowledge}, and 
(ii) prompting Large Language Models (LLMs) to identify associations between concepts \cite{Wadhwa2023RE-LLM, zhang2025kgnn}. 
SKT \cite{Tong2020SCKT} determines predecessor–successor and associative relations between KCs by analyzing the temporal order of interaction sequences. 
NGFKT \cite{Li2023NGFKT} calibrates the skill matrix and the Q-matrix via importance ranking of knowledge relations, accurately identifying KC relationships. 
SINKT \cite{Wadhwa2023RE-LLM} leverages LLMs' built-in knowledge and in-context learning to assist in constructing a KC relation graph that includes prerequisite concepts.

Despite progress in recent years, current KC-relation modeling remains incomplete and brittle. 
First, methods that infer relationships from interaction sequences rely on shallow cues, yielding sparse structures that overlook relations such as equivalence and containment and introducing erroneous relationships that conflate associative links with prerequisite–successor relations; 
prompt-based LLM approaches also suffer from generative uncertainty, leading to incorrect KC links. 
Such errors propagate through message passing and distort estimates of students' knowledge states. 
Second, many graph-based models encode the entire knowledge graph for each instance \cite{Li2025STHKT, Duan2024FusingRelationsDKT}, causing attention to diffuse into student-irrelevant regions, which amplifies noise and inflates computational cost. 
Consequently, graph-based KT methods take a double hit, noisy relations and inefficient computation, that undermines the reliability of the resulting knowledge graph. 
This motivates approaches that improve relational accuracy while concentrating computation on higher-value information.

\begin{figure*}[!th]
    \centering
    \includegraphics[width=0.95\linewidth]{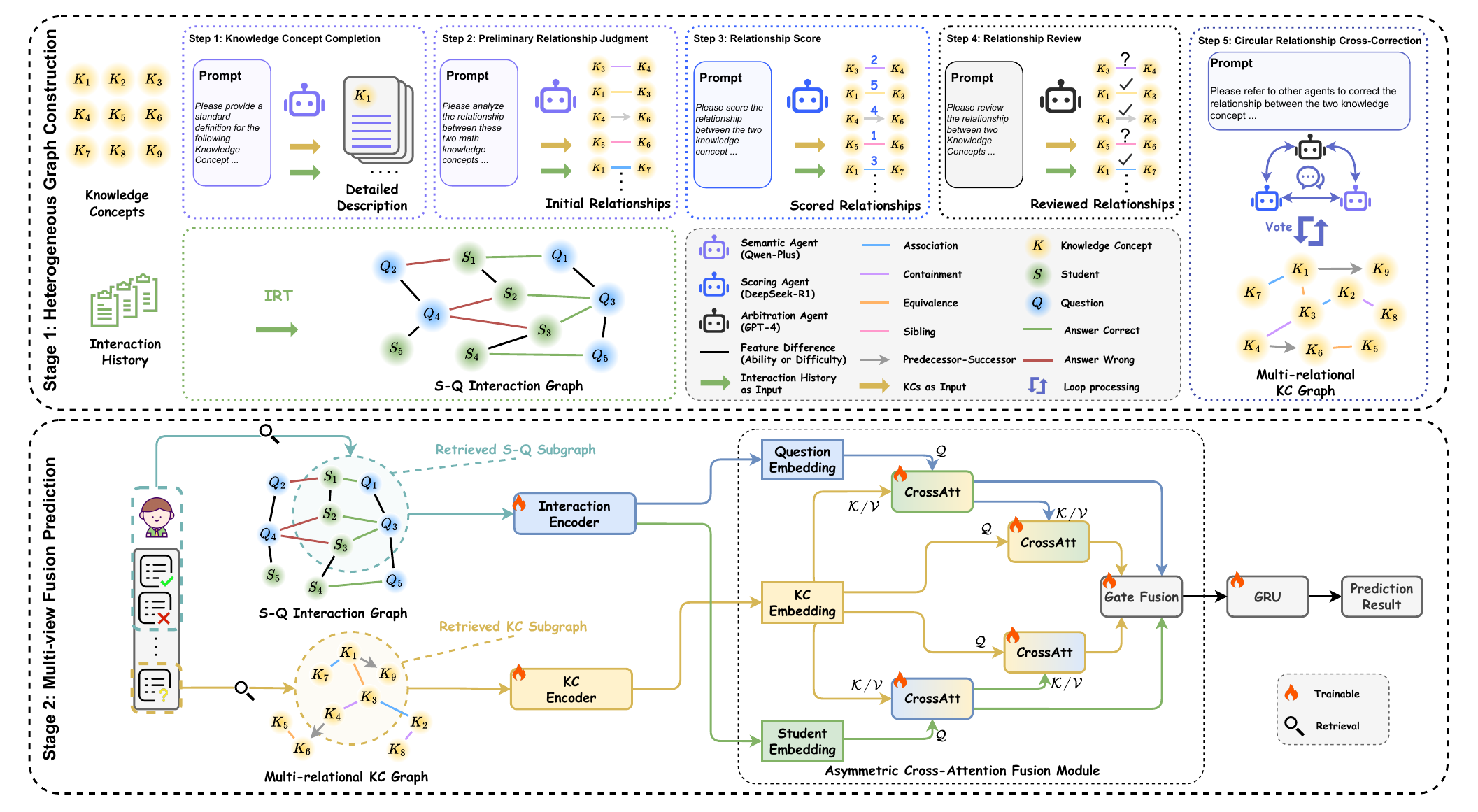}
    \vspace{-1em}
    \caption{The MAGE-KT framework.}
    \vspace{-1em}
    \label{fig:fig1}
\end{figure*}

To address these challenges, we propose MAGE-KT, a framework that couples relation quality with efficient computation through three mutually reinforcing designs. 
First, we construct a heterogeneous graph construction pipeline that integrates a multi-agent KC relation extraction module and a student–question interaction graph. 
The KC graph is enriched through a collaborative multi-agent pipeline, where specialized agents generate, score, and adjudicate five types of inter-KC relations, ensuring both semantic validity and structural consistency. 
Meanwhile, the S-Q interaction graph incorporates IRT-derived abilities and difficulties to model personalized student–question dynamics. The two graphs capture complementary dimensions of knowledge and learning behavior. 
Second, a student-conditioned subgraph retriever leverages the target student's history to jointly select high-value subgraphs from both views, focusing computation where it matters and avoiding irrelevant attention diffusion. 
Finally, we introduce an Asymmetric Cross-attention Fusion Module to fully capture the roles and interdependencies among students, questions, and knowledge concepts during prediction.
The main contributions of this paper are:
\begin{itemize}
    \item We propose MEGA-KT, a method that integrates a multi-agent–enhanced, multi-relational KC graph into multi-view KT for more accurate and efficient knowledge tracing. 
    \item We design a unified architecture that combines multi-agent KC graph construction, student–question interaction modeling, and asymmetric multi-view fusion, enabling accurate relation representation and efficient subgraph-based reasoning.
    \item We conduct comprehensive experiments on three widely used KT datasets, demonstrating that MEGA-KT achieves superior predictive accuracy, and that the multi-agent setting significantly improves KC-relation extraction accuracy.
\end{itemize}

\section{Methodology}
\label{sec:method}

\subsection{Heterogeneous Graph Construction}
This stage is as shown in the upper part of Fig.~\ref{fig:fig1}. 
The goal of this stage is to construct two complementary graphs, a Multi-relational KC Graph $\mathcal{G}_{laxK}$ and a Student–Question (S–Q) Interaction Graph $\mathcal{G}_{SQ}$, that supply inter-KC relations and calibrated ability/difficulty signals for subsequent student-conditioned retrieval and fusion.

\subsubsection{Multi-relational KC Graph.}
Let $\mathcal{K}=\{k_1,\ldots,k_{|\mathcal{K}|}\}$ be the set of KCs. 
We construct Multi-relational KC Graph $\mathcal{G}_{K}=(\mathcal{V}_{K},\mathcal{E}_{K},\tau)$, where $\mathcal{V}_{K}=\mathcal{K}$, $\mathcal{E}_{K}\subseteq \mathcal{V}_{K}\times\mathcal{V}_{K}$ denotes the set of KC–KC edges, and $\tau$ is the edge type. 
We consider five typed relations: 
(i) Association \cite{Gao2021RCD}: semantically related KCs that often co-occur without a strict teaching order; 
(ii) Containment \cite{Nakagawa2019GKT}: a part–whole or subconcept relation where $B$ is a component of $A$; 
(iii) Equivalence \cite{Jiang2016OntologyMatching}: two surface forms for the same concept; 
(iv) Sibling \cite{Tong2022ProblemSchema}: parallel subtopics under a common parent with no direct dependency; 
(v) Predecessor–Successor \cite{Pan2017PrereqMOOCs}: a prerequisite relation where $A$ must be mastered before $B$.

We employ a role-structured, three-agent pipeline to automatically generate and adjudicate five types of KC–KC relations. The collaborating agents include a Semantic Agent, a Scoring Agent, and an Arbitration Agent. The pipeline comprises five steps: 

\textbf{Step 1: Knowledge Concept Completion.} 
The Semantic Agent first standardizes KC names and performs semantic expansion, turning short labels into normalized descriptions that include definitions. The agent is given a prompt such as:
\par \textit{Please provide a standard 'definition' for the following knowledge concepts and determine their respective 'category.}

\textbf{Step 2: Preliminary Relationship Judgment.} 
Using the detailed KC descriptions together with evidence from the interaction history, the Semantic Agent selects the most plausible relation type from the five candidates for each KC pair, and outputs a structured record with evidence excerpts, source pointers, and key justifications.
We prompt the agent as follows:
\par \textit{Please analyze the structural relationship between these two math knowledge concepts based on the following content: definition, classification, precedence probability, and co-occurrence frequency, and judge their relationship type.}


\textbf{Step 3: Relationship Score.} 
Ingesting the Semantic Agent's structured output and interaction evidence, the Scoring Agent evaluates each candidate with type-specific criteria on a 0–5 scale. 
The instruction provided to the agent can be exemplified as follows:
\par \textit{Please score the "Predecessor-Successor" relationship between the two knowledge concepts based on the following criteria: Predecessor Dependency, Correctness Dependency, Answer Order Sequence. Please score each criterion (0–5) and provide a brief explanation for your scoring.}

\textbf{Step 4: Relationship Review.} 
The Arbitration Agent cross-checks semantic candidates, evidence, score vectors, and consistency tags, and validates them against type-level axioms and graph-topology constraints.
To guide this step, the agent is given a representative prompt such as:
\par \textit{Please based on the reasoning from Semantic Agent's recommendation, Scoring Agent's scoring, and the following definitions, classifications, and statistical information of the knowledge concepts, please determine the final relationship between Knowledge Concept A and Knowledge Concept B.}

\textbf{Step 5: Circular Relationship Cross-Correction.} 
Before and after arbitration, doubtful cases undergo two rounds of re-examination and voting. 
Using the structured evidence from the semantic and scoring stages, we also invoke three expert personas—teaching expert (checks prerequisite order and hierarchy against curricula), structure/semantics expert (tests entailment, equivalence, and sibling relations via definitions/terminology and parent–child links), and behavior/cognition expert (assesses data consistency via co-occurrence, temporal order, and performance dependence). 
Round 1 is a blind review in which the three experts independently return a relation type and justification. 
Round 2 is a reassessment with access to peer summaries, allowing revisions. 
For illustration, a simplified prompt is:
\par \textit{Please re-evaluating the relationship between A and B. Please judge the relationship based on the definitions, structure, and structured outputs from the multi-agent system, and provide a brief explanation.}

\subsubsection{S–Q Interaction Graph.}
Let $\mathcal{S}$ and $\mathcal{Q}$ denote students and questions, and let $\mathcal{D}=\{(s,q,r,t)\}$ denote the interaction log with correctness $r\in\{0,1\}$. 
We fit an IRT model to estimate student ability $\theta_s$ and question difficulty $b_q$, and attach them as node attributes. 
We then form Student–Question (S–Q) Interaction Graph $\mathcal{G}_{SQ}=(\mathcal{V}_{S}\!\cup\!\mathcal{V}_{Q},\mathcal{E}_{QS},\mathcal{E}_{QQ},\mathcal{E}_{SS})$, where $\mathcal{E}_{QS}$ connects $q$ and $s$ with label $r$, $\mathcal{E}_{QQ}$ links questions with weights $w_{qq'}=\exp(-|b_q-b_{q'}|/\sigma_q)$, and $\mathcal{E}_{SS}$ links students with weights $w_{ss'}=\exp(-|\theta_s-\theta_{s'}|/\sigma_s)$. 
These relationships support later student-conditioned retrieval of compact, high-value subgraphs while preserving the ability–difficulty structure for fusion.

\subsection{Multi-view Fusion Prediction}
This stage is as shown in the lower part of Fig.~\ref{fig:fig1}. 
The objective of this stage is to achieve efficient and accurate prediction. 
\subsubsection{Subgraph Retrieval}
Each prediction instance consists of a target question $Q_{\mathrm{tgt}}$ and the target student's historical interaction sequence: 
\begin{equation}
S_{\mathrm{tgt}} = \{(q_1, r_1), (q_2, r_2), \dots, (q_n, r_n)\},\quad r_i \in \{0,1\}
\end{equation}
IRT is used to evaluate the target student's ability and the difficulty of all questions $\{q_i\}^n_{i=1}$ appearing in the sequence. 
For the S-Q view, using the estimated ability and the difficulties of $\{q_i\}_{i=1}^n$, we locate in $\mathcal{G}_{SQ}$ the target student's node and the $n$ question nodes corresponding to $\{q_i\}_{i=1}^n$. 
These nodes serve as seeds; for each seed, we collect its $k$-hop neighborhood and then merge the neighborhoods to obtain the Retrieved S-Q Subgraph $\tilde{\mathcal{G}}_{SQ}$. 
This subgraph concentrates the evidence most relevant to predicting the target student's performance on the next question. 
For KC view, given the KC associated with $Q_{\mathrm{tgt}}$, we find the corresponding nodes in $\mathcal{G}_{K}$ and include all nodes within $n$ hops to form the Retrieved KC Subgraph $\tilde{\mathcal{G}}_{KC}$. 
This subgraph gathers concepts related to the target question's KC. 
Limiting the hop number bounds the influence of distant concepts and keeps the context focused.

\subsubsection{Asymmetric Cross-Attention Fusion}
We design an Asymmetric Cross-attention Fusion Module with two directional pathways (K–Q and K–S). 
Let the base embeddings for the concept, question, and student be $\mathbf{k}$, $\mathbf{q}$, and $\mathbf{s}$, respectively. 
First, inject question and student signals into the concept embeddings:
\begin{equation}
\begin{cases}
\widetilde{\mathbf{k}}^{(s)}=\mathrm{CrossAtt}(Q{=}\mathbf{k},\,K/V{=}\mathbf{s})\\[2pt]
\widetilde{\mathbf{k}}^{(q)}=\mathrm{CrossAtt}(Q{=}\mathbf{k},\,K/V{=}\mathbf{q})
\end{cases}
\end{equation}
Next, the enhanced concept evidence is propagated back to the student and question embeddings:
\begin{equation}
\begin{cases}
\widetilde{\mathbf{s}}^{(k)}=\mathrm{CrossAtt}(Q{=}\mathbf{s},\,K/V{=}\widetilde{\mathbf{k}}^{(s)})\\[2pt]
\widetilde{\mathbf{q}}^{(k)}=\mathrm{CrossAtt}(Q{=}\mathbf{q},\,K/V{=}\widetilde{\mathbf{k}}^{(q)})
\end{cases}
\end{equation}
To fuse the four streams, GateFusion first computes a gated sum of the two concept-enhanced paths and then concatenates this result with the student- and question-enhanced paths:
\begin{equation}
\begin{aligned}
\mathbf{z} 
&= GateFusion(\widetilde{\mathbf{k}}^{(s)}, \widetilde{\mathbf{k}}^{(q)}, \widetilde{\mathbf{s}}^{(k)}, \widetilde{\mathbf{q}}^{(k)}) \\&= 
\big[\widetilde{\mathbf{s}}^{(k)};\,\,(\alpha \cdot \widetilde{\mathbf{k}}^{(s)} + (1 - \alpha) \cdot \widetilde{\mathbf{k}}^{(q)});\,\widetilde{\mathbf{q}}^{(k)}\,\big]
\end{aligned}
\end{equation}
where $[\,\cdot\,;\,\cdot\,;\,\cdot\,]$ denotes concatenation. 

\subsubsection{Prediction and Loss Function}
A GRU maintains the student's evolving state and produces the next-question prediction from the fused vector:
\begin{equation}
\mathbf{h}_{t+1}=\mathrm{GRU}(\mathbf{h}_{t},\,\mathbf{z}),\qquad
\hat{y}_{t+1}=\sigma\!\left(\mathbf{w}_o^\top \mathbf{h}_{t+1}+b_o\right)
\end{equation}
where $\sigma(\cdot)$ is the sigmoid function and $\mathbf{w}_o,b_o$ are trainable parameters. To train all parameters, we use the Cross-Entropy Log loss between the predicted response $\hat{y}_{t+1}$ and the ground-truth response $r_{t+1}$ as the objective:
\begin{equation}
\mathcal{L}
= - \sum_{t=1}^{T-1} \Big( r_{t+1}\log \hat{y}_{t+1} + (1-r_{t+1})\log(1-\hat{y}_{t+1}) \Big)
\end{equation}

This two-stage, direction-sensitive design strengthens KC grounding for the current instance while avoiding unnecessary attention spread.

\section{Experiments and results}
\label{sec:experiment}
\subsection{Datasets}
We evaluate MAGE-KT on three public KT datasets: \textbf{ASSIST09} \cite{assessment}, \textbf{Junyi} \cite{junyi}, and \textbf{Statics2011} \cite{statics2011}.
We drop records with missing fields (student ID, correctness, or KC label), remove students with fewer than 10 attempts and questions answered fewer than 10 times, and perform student-level 8:1:1 splits (train/val/test) shared across methods.

\subsection{Baselines}
We group baselines into three categories: (i) DL-based Methods: DKT \cite{Piech2015DKT},
DKT+ \cite{Yeung2018PCR}, DKVMN \cite{Zhang2017DKVMN}; (ii) Transformer-based Methods: 
SAKT \cite{Pandey2019SAKT},
AKT \cite{Ghosh2020AKT}, SAINT \cite{Choi2020SAINT}; and (iii) Graph-based Methods: 
GKT \cite{Nakagawa2019GKT}, GIKT \cite{Yang2020GIKT}, MGEKT \cite{Qiu2024MGEKT}, TCL4KT \cite{Sun2023TCL4KT}, DyGKT \cite{Cheng2024DyGKT}, STHKT \cite{Li2025STHKT}, DGEKT \cite{Cui2024DGEKT}. These categories correspond to pure sequence modeling, attention-based modeling, and explicit structure modeling, respectively, enabling a comprehensive comparison against our method.

\subsection{Implementation Details}
We represent students, questions, and KCs using 128-dimensional learnable embeddings. Sequence modeling is handled by a GRU with hidden size 512 and dropout rate 0.3. Sequences longer than 100 are split into non-overlapping windows of length 100.
For multi-view fusion, we stack 3 layers of asymmetric cross-attention modules, implemented as pre-norm Transformer blocks (4 heads, model dimension 128), each followed by residual connections and feed-forward layers.
Training uses the Adam optimizer (lr 1e-3, weight decay 1e-5), batch size 64, and early stopping on validation AUC with patience of 10, for up to 100 epochs. Results are averaged over 3 random seeds.
In our multi-agent relation extraction module, we use Qwen-Plus \cite{qwen} as the Semantic Agent, DeepSeek-R1 \cite{deepseek} as the Scoring Agent, and GPT-4 \cite{gpt} as the Arbitration Agent.

\subsection{Main Results}

Table~\ref{tab:tab1} reports overall results on the three benchmarks. 
On ASSIST09, Junyi, and Statics2011, MAGE-KT attains the best AUC and ACC, indicating robust generalization. 
Unlike purely sequential DL baselines, MAGE-KT encodes students, questions, and KCs into embeddings and performs directed information fusion via an asymmetric cross-attention module, addressing the difficulty of capturing long-range dependencies and structural relations. 
Compared with Transformer baselines, MAGE-KT's directed information flow suppresses redundant attention and ineffective interactions, yielding more stable performance with reduced overfitting across data distributions and scales. 
Relative to graph-based methods, MAGE-KT reduces reliance on deep graph propagation by operating on retrieved subgraphs and using lightweight encoders plus directed fusion, improving predictive accuracy and interpretability. 
These results validate the effectiveness of MAGE-KT for KT.

\begin{table}[t]
\centering
\setlength{\tabcolsep}{6pt}
\caption{Main results on three datasets. Underlined values indicate the second-best.}
\label{tab:tab1}
\scalebox{0.74}{
\begin{tabular}{@{}ccrrrrrr@{}}
\toprule
\multirow{2}{*}{\textbf{Categories}} & \multirow{2}{*}{\textbf{Baselines}} & \multicolumn{2}{c}{\textbf{ASSIST09}} & \multicolumn{2}{c}{\textbf{Junyi}} & \multicolumn{2}{c}{\textbf{Statics2011}} \\ \cmidrule(lr){3-4}\cmidrule(lr){5-6}\cmidrule(lr){7-8}
 &  & \multicolumn{1}{c}{\textbf{ACC}} & \multicolumn{1}{c}{\textbf{AUC}} & \multicolumn{1}{c}{\textbf{ACC}} & \multicolumn{1}{c}{\textbf{AUC}} & \multicolumn{1}{c}{\textbf{ACC}} & \multicolumn{1}{c}{\textbf{AUC}} \\ \midrule
\multirow{3}{*}{\textbf{\begin{tabular}[c]{@{}c@{}}DL-based \\ Methods\end{tabular}}} & DKT & 74.86 & 78.34 & 81.89 & 82.52 & 79.72 & 81.03 \\
 & DKT+ & 75.02 & 79.84 & 81.77 & 82.98 & 79.77 & 82.22 \\
 & DKVMN & 74.44 & 78.97 & 81.41 & 83.89 & 78.29 & 79.93 \\ \midrule
\multirow{3}{*}{\textbf{\begin{tabular}[c]{@{}c@{}}Transformer-based\\ Methods\end{tabular}}} & SAKT & 74.56 & 77.98 & 79.81 & 81.08 & 79.62 & 79.17 \\
 & AKT & 76.47 & 81.61 & 83.42 & 84.94 & 79.87 & 84.21 \\
 & SAINT & 71.86 & 75.58 & 84.97 & 85.78 & 79.82 & 79.99 \\ \midrule
\multirow{7}{*}{\textbf{\begin{tabular}[c]{@{}c@{}}Graph-based\\ Methods\end{tabular}}} & GKT & 75.37 & 78.98 & 79.72 & 84.86 & 79.02 & 80.40 \\
 & GIKT & 75.81 & 80.63 & 84.37 & 86.38 & \underline{82.75} & 86.07 \\
 & MGEKT & 79.32 & 86.12 & 87.89 & 88.89 & 81.61 & 85.39 \\
 & TCL4KT & 76.26 & 81.58 & 85.68 & 87.44 & 82.56 & \underline{86.81} \\
 & DyGKT & 78.29 & \underline{86.67} & \underline{88.02} & \underline{89.62} & 83.27 & 84.36 \\
 & STHKT & \underline{79.53} & 84.72 & 81.29 & 86.88 & 80.79 & 82.58 \\
 & DGEKT & 77.12 & 81.56 & 79.82 & 81.26 & 81.97 & 82.85 \\ \midrule
\textbf{Ours} & \textbf{MAGE-KT} & \textbf{83.06} & \textbf{87.89} & \textbf{90.33} & \textbf{91.79} & \textbf{87.29} & \textbf{87.72} \\ \bottomrule
\end{tabular}
}
\end{table}

\subsection{Ablation Results}
\subsubsection{Model Components.}
Table~\ref{tab:tab2} reports ablations over the main modules and shows that removing any component consistently lowers ACC/AUC on all three datasets, confirming both the necessity and complementarity of MAGE-KT's designs. 
\textbf{w/o AsyAtt}: dropping the asymmetric cross-attention degrades performance, indicating that role-aware, directed fusion is crucial for aligning student/question evidence with KC semantics and for suppressing redundant interactions. 
\textbf{w/o KC Graph}: removing explicit KC structure forces the model to rely mainly on local sequential statistics, weakening cross-item transfer and concept-level inference. 
\textbf{w/o S-Q Graph}: this yields the largest drop, showing that student–question structure (ability/difficulty and temporal context) is indispensable for capturing individual trajectories and item heterogeneity. 
\textbf{w/o Subgraph}: disabling student-conditioned subgraph retrieval harms accuracy, validating that focusing computation on compact, high-value neighborhoods reduces noise and attention diffusion. 
Overall, the ablations substantiate MAGE-KT's design choices: explicit structure (KC and S-Q graphs), targeted retrieval, and directed fusion act synergistically to improve next-question prediction.

\begin{table}[t]
\centering
\caption{Component ablations on three datasets.} 
\label{tab:tab2}
\setlength{\tabcolsep}{5pt}
\scalebox{0.9}{
    \begin{tabular}{lcccccc}
    \hline
    \multirow{2}{*}{Methods} & \multicolumn{2}{c}{ASSIST09} & \multicolumn{2}{c}{Junyi} & \multicolumn{2}{c}{Statics2011} \\
    \cmidrule(lr){2-3}\cmidrule(lr){4-5}\cmidrule(lr){6-7}
     & ACC & AUC & ACC & AUC & ACC & AUC \\
    \hline
    Full          & \textbf{83.06} & \textbf{87.89} & \textbf{90.33} & \textbf{91.79} & \textbf{87.29} & \textbf{87.72} \\
    w/o AsyAtt    & 79.87 & 83.84 & \underline{84.77} & \underline{85.86} & 82.34 & 83.82 \\
    w/o KC Graph  & 81.57 & 85.25 & 83.75 & 85.23 & \underline{84.45} & 84.76 \\
    w/o S-Q Graph & 78.51 & 80.39 & 80.81 & 83.86 & 81.71 & 82.94 \\
    w/o Subgraph  & \underline{81.65} & \underline{85.69} & 84.76 & 85.56 & 84.35 & \underline{85.93} \\
    \hline
    \end{tabular}%
}
\vspace{-1em}
\end{table}

\subsubsection{Prerequisite–Successor Extraction (Junyi).}
To validate the effectiveness of our Multi-agent Relationship Extraction pipeline, we conduct experiments on the Junyi dataset, which provides curated prerequisite–successor relations. 
As shown in Table~\ref{tab:tab3},  the Full pipeline achieves the highest prediction (Pred), correctness (Corr), and Jaccard (Jacc), outperforming the single-agent variant (Qwen-Plus only) and the ablations without Completion or Correction. 
Removing Completion mainly hurts Pred and Jacc, indicating reduced recall when KC descriptions are not enriched. 
Removing Correction primarily lowers Corr, reflecting unresolved spurious or misdirected relationships. 
These results confirm that the Multi-agent Relationship Extraction pipeline yields a higher-fidelity concept graph that benefits downstream KT.

\begin{table}[t]
\centering
\caption{Ablation experiment of Prerequisite-Successor Extraction on Junyi.} 
\label{tab:tab3}
\setlength{\tabcolsep}{6pt}
\scalebox{0.85}{
    \begin{tabular}{lccc}
    \hline
   {Method} & {Pred (\%)} & {Corr (\%)} & {Jacc (\%)} \\
    \hline
        Full & \textbf{92.52} & \textbf{91.73} & \textbf{85.46} \\ 
        Single-agent (Qwen-Plus only) & 79.31 & 78.64 &  \underline{77.11} \\
        w/o Completion & 83.40 & 81.57 & 70.23 \\
        w/o Correction &  \underline{85.92} & \underline {85.61} & 75.10 \\
    \hline
    \end{tabular}%
}
\end{table}

\section{Conclusion And Future Work}
In this paper, we present MAGE-KT, a multi-agent, multi-view framework with asymmetric cross-attention for knowledge tracing. 
It coordinates heterogeneous agents to produce higher-quality inter-KC relations. 
Conditioned on the target student's interaction history, it retrieves compact, high-value subgraphs and fuses them via the asymmetric module. 
MAGE-KT achieves state-of-the-art results on multiple benchmarks and consistently surpasses sequence-based and full-graph baselines in accuracy.
In the future, we plan to extend MAGE-KT to continual settings for real-time adaptation, incorporate richer textual signals to refine KC relations, and study uncertainty calibration and human-in-the-loop correction to further improve reliability in deployment.
\label{sec:conclusion}

\section{Acknowledgments}
This work was funded by the National Natural Science Foundation of China (No. 62567005), and Natural Science Foundation of Inner Mongolia Autonomous Region of China (No. 2025MS06004).
\bibliographystyle{IEEEbib}
\bibliography{refs}

\end{document}